\documentclass[conference]{IEEEtran}
\IEEEoverridecommandlockouts

\usepackage{cite}
\usepackage{amsmath,amssymb,amsfonts}
\usepackage{graphicx}
\usepackage{textcomp}
\usepackage{xcolor}
\def\BibTeX{{\rm B\kern-.05em{\sc i\kern-.025em b}\kern-.08em
    T\kern-.1667em\lower.7ex\hbox{E}\kern-.125emX}}
\usepackage[colorinlistoftodos]{todonotes}
\presetkeys{todonotes}{inline}{}
\usepackage{algorithm, float, listings}
\usepackage[noend]{algpseudocode}
\usepackage{csquotes}
\usepackage{graphicx}
\usepackage{multirow}
\usepackage{subcaption}
\usepackage{wrapfig}
\usepackage{subcaption}
\usepackage{amsmath}
\usepackage{arydshln}
\usepackage{booktabs}
\usepackage{hyperref}

\newcommand{\merge}{Retrieve and Merge}
\newcommand{\extractdots}{Condense and Extract Dots}
\newcommand{\db}{Save into Database}

\begin{document}
\bstctlcite{IEEEexample:BSTcontrol}
\IEEEpubid{\makebox[\columnwidth]{Accepted in IEEE Big Data 2024. \copyright\ 2024 IEEE \hfill}
\hspace{\columnsep}\makebox[\columnwidth]{ }}

\title{LLM Augmentations to support \\Analytical
Reasoning over Multiple Documents}

\author{\IEEEauthorblockN{Raquib Bin Yousuf, Nicholas Defelice, Mandar Sharma, Shengzhe Xu, Naren Ramakrishnan} \IEEEauthorblockA{Department of Computer Science, Virginia Tech, Arlington, VA\\
Email: raquib@vt.edu, naren@cs.vt.edu}}

\maketitle

\begin{abstract}
Building on their demonstrated ability to perform a variety of tasks, we investigate the application of large language models (LLMs) to enhance in-depth analytical reasoning within the context of intelligence analysis. Intelligence analysts typically work with massive dossiers to draw connections between seemingly unrelated entities, and uncover adversaries' plans and motives. We explore if and how LLMs can be helpful to analysts for this task and develop an architecture to augment the capabilities of an LLM with a memory module called dynamic evidence trees (DETs) to develop and track multiple investigation threads. Through extensive experiments on multiple datasets, we highlight how LLMs, as-is, are still inadequate to support intelligence analysts and offer recommendations to improve LLMs for such intricate reasoning applications.
\end{abstract}

\begin{IEEEkeywords}
augmented LLM, intelligence analysis, analytical reasoning, retrieval and search, large language model.
\end{IEEEkeywords}


\section{Introduction}
The impressive generation capabilities of LLMs~\cite{kojima2022large, wei2022chain, bubeck2023sparks}, 
along with their successful application in diverse areas~\cite{kumar2023large, kaddour2023challenges, medical_education, dtg_survey}
led us to explore their utility for intelligence analysis (IA).
Key IA tasks involve uncovering plots from textual reports such that interventions can be made to prevent unfortunate events. This entails making connections between seemingly unrelated entities and events. This undertaking traditionally requires significant investments of time and effort from human analysts~\cite{kang_characterizing, davidson2022exploring}. Essentially, they are carrying out
a ``connecting the dots'' task, where dots are the information bits involving different entities/events in their reports.
\begin{figure}[h]
  \centering
    \includegraphics[width=\columnwidth]{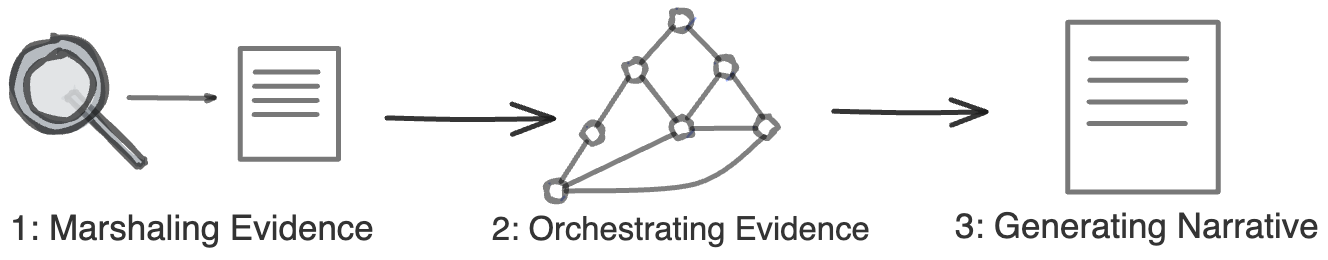}
  \caption{Three steps to intelligence analysis (IA).}
  \label{fig:steps}
\end{figure}
\vspace{-0.2cm}

IA can be viewed as comprising
the following subtasks: 
i) Marshaling evidence, 
ii) Orchestration of the gathered evidence: sensemaking and the construction of defensible and persuasive arguments from evidence, and 
iii) narrative generation. 
To be successful in IA,
analysts often require an overarching capability of creative, speculative, and imaginative reasoning which helps build hypotheses. Even after making the connection between relevant information `dots', an analyst must decide what this synthesized information portrays for the context at hand. New evidence can support or topple the hypothesis under consideration. After gathering all required evidence, analysts need to produce convincing arguments and reasoning to generate an appropriate alert to their superiors/authorities.

While numerous representations have been proposed for reasoning
with LLMs~\cite{touvron2023llama, touvron2023llama2, ouyang2022instructgpt} they
are typically focused on individual questions or tasks~\cite{wei2022chain, yao2024tree, besta2024graph}.
Our focus here is on reasoning emergent from
assimilating hundreds of documents,
sometimes beyond context window limits for LLMs.
We design a set of experiments to 
study whether LLMs can solve
IA problems and if not, how can they be augmented to support such analysis. 
Our codebase is publicly available at \href{https://github.com/DiscoveryAnalyticsCenter/speculatores}{https://github.com/DiscoveryAnalyticsCenter/speculatores}.
Our key contributions are:
\begin{enumerate}
\item We conduct the first investigation of the feasibility of using LLMs in intelligence analysis
where both evidence-based
reasoning and analytical creativity is of utmost importance.
\item We develop
a three-step augmentation to support
the use of LLMs for IA:
i) Dynamic Evidence Trees (DETs), a memory module to help organize evidences, ii) Data condensation via LLMs, and iii) an LLM-driven search and retrieval process.
\item Applying our framework on multiple IA datasets, we show that while our augmentations help orchestrate and
improves narratives on large datasets, LLMs still lack the analytical creativity to craft
convincing arguments.
\item We outline detailed
recommendations for applying LLMs to IA tasks and specifically how they can serve as modules in specific subtasks such as evidence marshalling and narrative generation.
\end{enumerate}
\section{Methodology}
\vspace{-0.1cm}
We design our experiments with three intelligence analysis datasets. These datasets contain field reports of various events conducted by adversaries. Among these, some may be relevant and others are irrelevant. Analysts need to find  pertinent information by isolating relevant reports and piecing the information together to speculate the main plot. We pose four research questions (RQ) to understand whether LLMs can be useful for the IA process:
\vspace{-0.1cm}
\begin{itemize}
    \item \textbf{RQ1:} \textit{Can LLMs solve IA problems on their own?}
    \item \textbf{RQ2:} \textit{Does augmentation help? If so what kinds of augmentation support the IA task?}
    \item \textbf{RQ3:} \textit{What is the effect of temperature and allowable context size on the  skills of LLMs?}
    \item \textbf{RQ4:} \textit{Where can LLMs contribute in the 3-step process outlined in section \textsection 1?}
\end{itemize}
\vspace{-0.1cm}
We start with an overview of how the experiments are designed to capture the efficacy of LLMs in IA. The goal of intelligence analysis is not just to summarize the reports, but rather to find out the connections, along with the respective backstories about the adversaries. A major limitation for LLMs to achieve this is the context length and loss of attention with the increasing input size \cite{levy2024same}.  
\vspace{-0.1cm}
\subsection{Preliminaries}
\vspace{-0.1cm}
\label{ssec:problem_definition}
Each IA dataset is a set of chronological reports $\mathcal{R}$ = $\{r_1, r_2, \dots,  r_N\}$ such that report $r_{i-1}$ originates or was assigned before $r_{i}$. When there are multiple reports available on a single day, we can order them arbitrarily. 

We also have a set of evidential information nuggets called `dots' $\mathcal{E_D}$ = $\{e_{d1}, e_{d2}, \dots,  e_{dm}\}$. We can model each report $r_i$ as one evidential dot $e_{dmi}$ or an amalgamation of  multiple evidential dots, i.e.,  $r_i$ = $\{e_{d1}, e_{d2}, \dots,  e_{dm}\}$. Thus, we can represent each dataset as a tuple of $(\mathcal{R},\mathcal{E_D})$. 

Conceptually there can be two types of information dots in IA: i) evidential dots $e_{di}$, which come directly from the intelligence reports; ii) hypothesis dots $h_{di}$, which are produced by synthesizing multiple evidential dots (and/or other hypothesis dots). Thus each hypothesis dot is modeled as $h_{di} = \{h_{d1}, h_{d2}, \dots,  h_{dk}\}, \dots \{e_{d1}, e_{d2}, \dots,  e_{dl}\}$. 

The goal is to use the tuple $(\mathcal{R},\mathcal{E_D})$ to create a set of hypothesis dots, i.e. $\mathcal{H_D} = \{{h_d}_1, {h_d}_2, \dots,  {h_d}_n\}$. In the most basic format, a hypothesis dot ${h_d}_i$ can thought to be built with two evidential dots; $h_{di} = \{e_{dq}, e_{dp}\}$. Thus, the LLM autoregressively generates each token  ${h}_{di, j}$ of a hypothesis dot $h_{di}$ as:
\vspace{-0.1cm}
\begin{align}
    \mathbb{P}(\textbf{h}_{di}) &= \mathbb{P}(h_{di,1}, h_{di,2}, ...,h_{di,n}) \\
    &\simeq \prod_{j=1}^{n} \mathbb{P}({h}_{di, j} | h_{di,1}, h_{di,2}, ...,h_{di,j-1})
\end{align}
We postulate that the autoregressive nature of LLMs will help model a report with evidential dot (or dots) as a whole, instead of modeling reports as an amalgamation of a set of entities. We note that each evidential dot $e_{di}$ is comprised of a set of entities. The autoregressive nature of LLMs will keep the rich contexts of evidential dots intact and build up each hypothesis dot ($h_{dj}$). We are postulating that LLMs with their auto-regressive and generalization properties will be able to model dots as standalone artifacts and connect them when needed.
\vspace{-0.1cm}
\begin{figure}[h]
  \centering
    \includegraphics[width=\columnwidth]{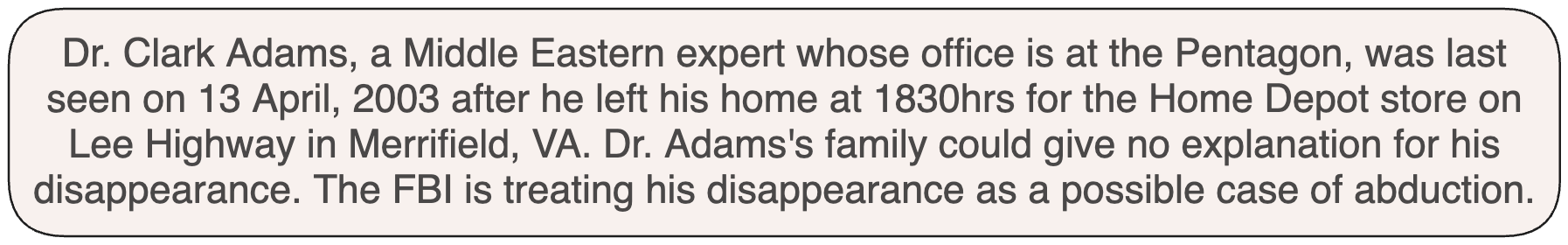}
  \caption{Sample intelligence report}
\end{figure}
\vspace{-0.5cm}
\subsection{Initial experiments}
\label{ssec:initial}
\vspace{-0.1cm}
As the most basic step of the evaluation of LLMs for IA, we attempt to use each dataset in its entirety in the context windows, as the limit permits, to generate the narrative. We empirically test with varied system prompts. However, we find that the LLMs tend to only summarize the documents and present superficial information, despite numerous prompt variations to bring out implicit connections and the broader story. Moreover, the context length limits the number of documents that can be fit into the prompt. We randomly sub-sample the number of documents for datasets according to the context limit. We also quantitatively and qualitatively evaluate the responses as shown in the results section \ref{sec:results}.

\subsection{LLMs for Intelligence Analysis: Challenges}
\label{ssec:llm_challenges}
Through our initial experiments on LLMs without any augmentations, we identified some shortcomings for complex analytical reasoning tasks. A major limitation for LLMs is the context length and loss of attention with the increasing input size \cite{levy2024same}. The two main challenges are i) lack of a proper memory module to keep track of all the evolving investigation threads, ii) limited context length that limits the number of reports that can be processed. To solve the first problem, we propose an augmented architecture with a much needed memory module to the pipeline, named ``Dynamic Evidence Trees (DETs)''. Memory modules are increasingly being used to augment LLMs in various applications \cite{park2023generative}. As such, we devise a memory module in form of trees to help LLMs orchestrate evidence as the reasoning goes along. As an improvement on the second front, we also perform the tests with two different granularities of the reports. We propose condensing the reports into concise information chunks. In the later sections, we will formally define and present the inner workings of these two components.
\begin{figure}[h]
\centering
    \begin{subfigure}[t]{0.90\columnwidth}
        \centering
        \includegraphics[width=\linewidth]{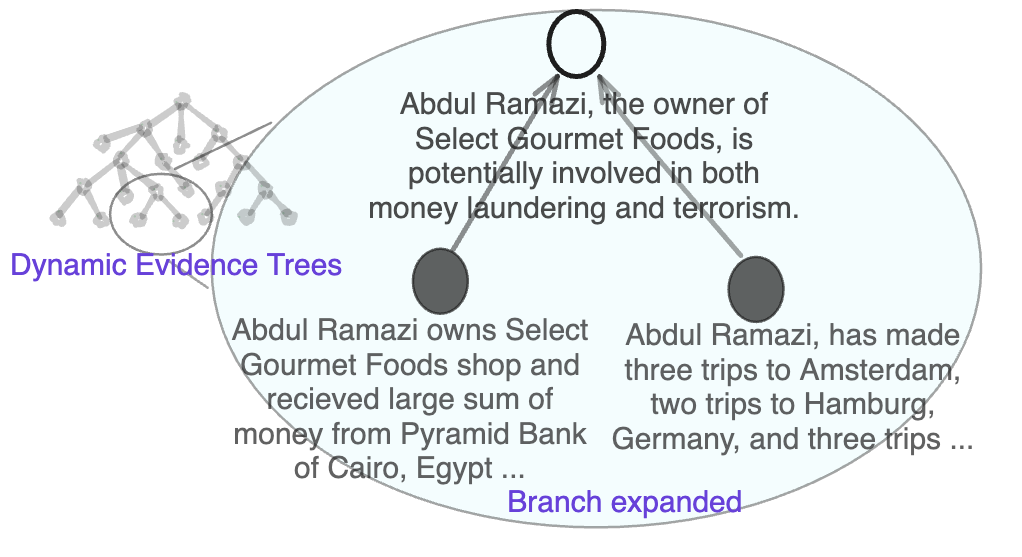}
        \caption{Regular DETs, dots: (black: document, white: hypothesis)}
    \end{subfigure}
    \begin{subfigure}[b]{0.90\columnwidth}
        \centering
        \includegraphics[width=\linewidth]{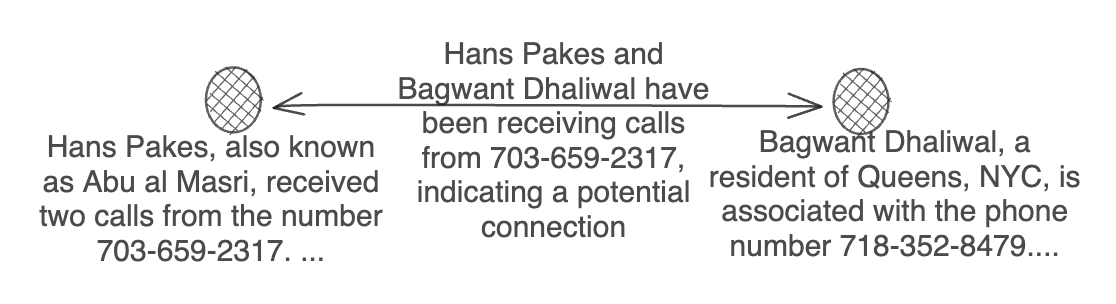}
        \caption{Person-based DETs (shaded dots)}
    \end{subfigure}
    \caption{Dynamic Evidence Trees}
\end{figure}
\vspace{-0.2cm}
\subsection{First augmentation: Dynamic Evidence Trees (DETs)}
\label{ssec:det}As an improvement to the basic LLM, we augment it with dynamic evidence trees (DETs). This is the main memory component to save the evidence. 

We define the main graph structure of DETs as $\mathcal{G}= (\mathcal{V}, \mathcal{E})$ where $\mathcal{V}$ denotes the vertices with $\mathit{v} \in \mathcal{V}$ such that each vertex $\mathbf{\mathit{v_i}}$ is a structure named DOT (i.e., information dot); and $\mathcal{E}$ denotes edges $\mathit{e_{ij}} \in \mathcal{E}$ which is the parent-children relationship between the $\mathit{DOT_i}$ and $\mathit{DOT_j}$. Thus set of vertices $\mathcal{V}$ can be defined as $\{DOT_1, DOT_{2}, \dots DOT_{i}\}$.
Each $\mathit{DOT_i}$ can be both evidential and hypothesis dots and we define $\mathit{DOT_i}$ as a quadruple of $(\mathit{information, DOT_{children}, DOT_{parents}, document})$, such that each $\mathit{DOT_i}$ can have another set dots as children or parents. Evidential dots belong to one report of the dataset and we save that information in the node. DETs also holds a database object with LLM based embedding vectors for retrieval operation. Thus we denote DETs as tuple of $(\mathcal{DB}, \mathcal{G})$. During the creation of the DETs, we can think of each $\mathit{DOT}$ as a subgraph with a tree like structure (with one node if it does not have any children or parents). LLM based operations will decide if multiple $\{DOT_1, DOT_{2}, \dots DOT_{i}\}$ can be merged to create a new hypothesis DOT. This new DOT will have all the comprising dots as its children and the children dots will also save the new DOT as their parents. Thus, we are creating new connections among different disjoint subgraphs $\{DOT_1, DOT_{2}, \dots DOT_{i}\}$ in the form of a new parent DOT. 

We also experiment with a variant of DETs, where each dot is amalgamation of all information of a person in the dataset and the edges are defined as connections between two persons. Thus we can define DETs as $\mathcal{G}= (\mathcal{V}, \mathcal{E})$ where $\mathcal{V}$ denotes the vertices with $\mathit{v} \in \mathcal{V}$ such that each vertex $\mathbf{\mathit{v_i}}$ is information about a person; and $\mathcal{E}$ denotes edges $\mathit{e_{ij}} \in \mathcal{E}$ which is the connections between two persons, i.e., $\mathit{DOT_i}$ and $\mathit{DOT_j}$.

With the help of DETs, LLMs try to build up hypotheses and investigation threads. Intelligence analysts go through a process of discovery and combining the dots to build up various hypotheses. Like human analysts might do, DETs helps to keep track of all the information dots and keeps connecting them with new relevant information in a tree like structure. Each hypothesis dot represents an investigation thread that may or may not get added to another related investigation thread. We report results for both regular and person-based DETs in section \ref{sec:results}
\begin{figure}[h]
  \centering
    \includegraphics[width=\columnwidth]{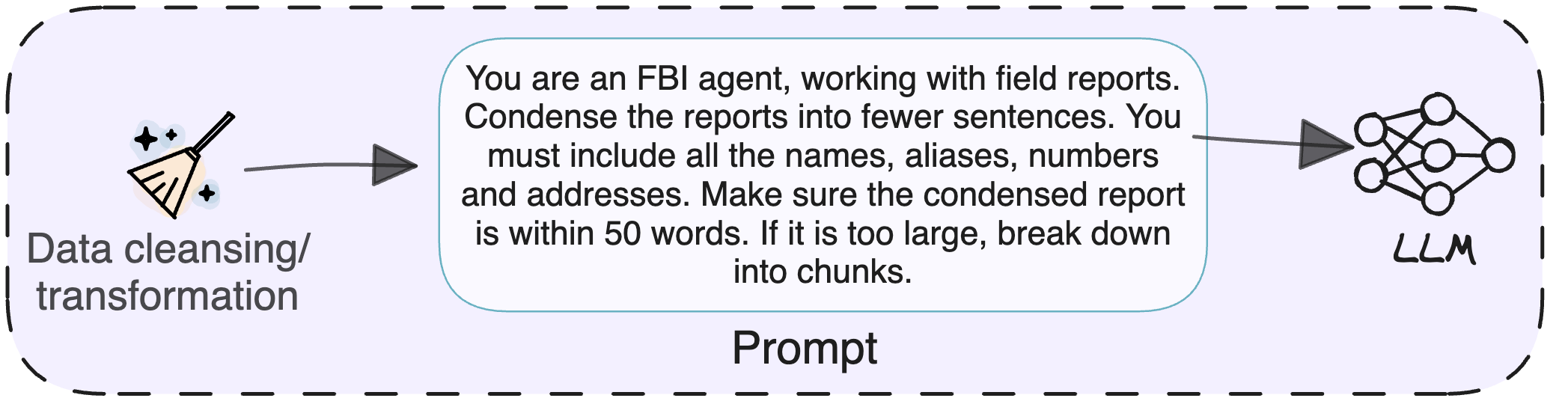}
  \caption{Data condensation and dot extraction.}
\end{figure}
\vspace{-0.5cm}
\subsection{Second augmentation: Data Condensation}
\label{ssec:condense}
We utilize the language modeling capabilities of LLMs to digest the set of reports into usable information dots before generating reports. These condensed information dots can be merged with other information dots to create hypotheses. We empirically test various system prompts to break down the report in a zero-shot fashion such that a report $r_{i}$ turns into an evidential dot ${e_d}_i$. We use dynamic pre-processing to break it down further, if deemed necessary by the LLM. All the reports in our experiments yielded a single dot each. This also indicates the concise reporting practice of the intelligence community. 
\begin{figure*}[t]
  \centering
    \includegraphics[width=0.95\textwidth]{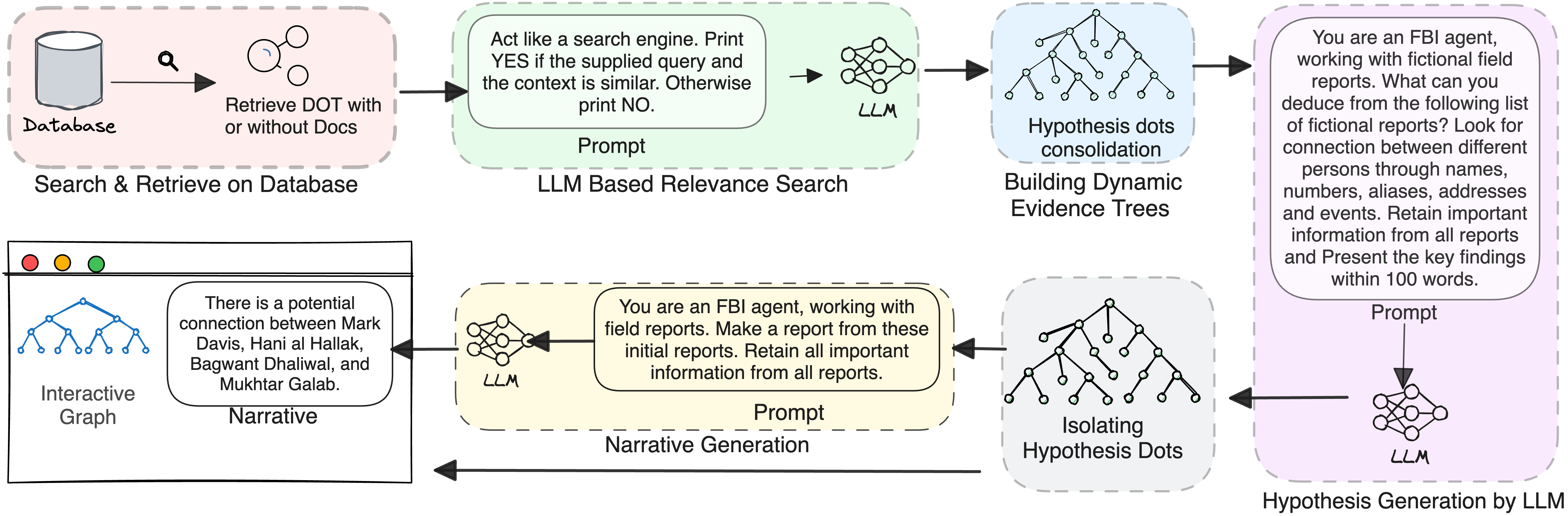}
  \caption{All augmentations together with retrieve and merge}
  \label{fig:all_together}
\end{figure*}
\begin{figure}[h]
\begin{minipage}{\columnwidth}
    \begin{algorithm}[H]
    \caption{Build and Update DETs}
    \label{algo:build_det}
    \scriptsize
    \label{algo:build_det}
    \begin{algorithmic}[1]
    \Require List of ordered reports, \(\mathcal{R}\)
    \Ensure Dynamic Evidence Trees, $\mathsf{DETs}$ $(\mathcal{DB}, \mathcal{G})$
    
    \Function{\extractdots}{$\mathit{r_1}$}
        \State Invoke $\mathcal{LLM(\mathit{r_1})}$
    \EndFunction
    \Function{\db}{DOT}
        \State Invoke Embedding $\mathcal{LLM}$
        \State Save Embedding to DB
    \EndFunction
    
    \For{\(r \in \mathcal{R}\)}
        \State $DOTs$ = \Call{Extract DOT}{$r$}
        \For{$DOT \in DOTs$}
            \State \Call{\merge\ref{algo:merge}}{$DOT$}
            \State \Call{\db}{$DOT$}
        \EndFor
    \EndFor
    \end{algorithmic}
    \end{algorithm}
\end{minipage}

\begin{minipage}{\columnwidth}
    \begin{algorithm}[H]
    \scriptsize
    \caption{\merge}
    \label{algo:merge}
    \begin{algorithmic}[1]
    \Require $\mathit{DOT}$
    
    \Function{Retrieve}{$DOT_{query}$}
        \State Vector Search in $DETs$
    \EndFunction
    
    \Function{LLM based Filtering}{$DOT_{candidates}$, $DOT_{query}$}
        \State Invoke $\mathcal{LLM}$
    \EndFunction
    
        \State $DOT_{candidates} \gets \Call{Retrieve}{DOT}$
        \State $DOT_{filtered} \gets \Call{LLM based Filtering}{DOT_{candidates}, DOT_{query}}$
        \If{$|DOT_{filtered}|\neq \emptyset$} 
            \State $DOTs_{candidates} \gets \Call{Hypo. Node Consolidation}{DOTs_{filtered}}$
            \State $DOTs_{candidates} \gets \Call{Evid. Node Collection}{DOTs_{candidates}}$
            \State $DOT_{new} \gets $ Invoke $\mathcal{LLM}$
            \State \Call{\merge\ref{algo:merge}}{$DOT_{new}$}
            \State \Call{\db}{$DOT_{new}$}
        \EndIf
            
    \end{algorithmic}
    \end{algorithm}
\end{minipage}%
\end{figure}
\subsection{Third augmentation: LLMs for Retrieval}
To build up a DET, we require search and retrieval capabilities on the existing evidence list, given new evidence. We augment the LLMs with a two-step retrieval pipeline. The first step is powered by an LLM-based embedding vector database and similarity-based search. Following the recent use of LLMs-based embeddings for retrieval \cite{muennighoff2022mteb, huang2020embedding}, we adopt an instructor embedding model \cite{su2023embedder} for our database. The second step uses an LLM to filter the extracted set of DOTs.
\subsection{The augmentations altogether}
\label{ssec:all_together}
The augmented version of the experiment is designed to process intelligence report sequentially and keep track of the evidence in DETs. 
This sequential approach helps in three ways; first, to conform to the dynamic nature of how intelligence report are available and assigned over period of time; second, to help build up DETs to keep track of the emerging evidence; third, to get around the context limitation of the models. This also follows the natural tendencies of how analyst must make and keep the hypotheses running until enough evidence can be accrued and a reasoning can be established for each hypothesis. New reports and the resulting information can either approve or disprove any existing hypotheses. The DET-based augmented pipeline considers this aspect and inputs the reports iteratively to simulate the temporal continuum of intelligence analysis. 

Each report goes through a pre-processing step based on the dataset provided and a set of information dots are extracted from the report. We initialize a database and keep updating it with new information dots during runtime. During the tree building operation, the system first attempts to search relevant DOTs from the DETs. 
The extracted $DOTs_{candidates}$ go through a parent level hypothesis DOTs identification and consolidation process. We take the lowest common parent hypothesis nodes and assign the new DOT to that node. This ensures that we are assigning the new DOT to the most relevant branch of the evidence tree. Afterwards, each evidential dots of the resulting new branch are extracted. LLMs will synthesize all the evidential dots to create a short narrative for this particular investigation thread.
From the resulting DETs, we isolate the largest chain of events and use that as the main DET for the input. The full generated DETs along with its disjoint nodes also demonstrate how LLMs are being augmented to keep track different documents and how each of these documents are being utilized through the chain of events for building up to the final hypothesis. Fig. \ref{fig:all_together} shows the final steps of the proposed architecture and algorithm \ref{algo:build_det} and \ref{algo:merge} shows the overall algorithm used for the build and merge operation.
\vspace{-0.1cm}
\section{Evaluation setup}
In this section, we describe the layout of our evaluation procedure. Because we aim to test the efficacy of LLMs as an intelligence analyst, we evaluate the augmented architecture in an ablated fashion. We remove augmentations one by one and capture the improvement from the most basic bare-bone LLM. We also consider document-entity networks and clustering as a more traditional baseline sans any language modeling. For the basic form, we adopt two versions, one with a sub-sampled report set and another with the highest performing clusters on the report set coming from one of the traditional baselines. We use default parameters for the generation, with an enumerative testing with temperature and word limit to capture the randomness and creativity in reasoning. 
Our architecture is designed to use any type of LLM, both through API and local storage. It can fall back to a local LLM if it fails to get a results from the API calls. 
We experimented with five models from four different model families: i) GPT-3.5, and ii) GPT-4 from OpenAI, iii) Llama-2 \cite{touvron2023llama2} from Meta AI, iv) Mistral-7B \cite{jiang2023mistral7b} from MistralAI, and v) Gemma-2 \cite{gemma} from Gemini platform, Google.
\vspace{-0.2cm}
\subsection{Datasets and ground truth}
\label{ssec:dataset}
We utilize three datasets---Sign of the Crescent(Crescent), Atlantic Storm and Manpad---popular in training and analytics competitions~\cite{wu2012start}. 
The Crescent and Atlantic Storm datasets have their solutions divided into a few subplots and charts. Crescent has 3 subplots, each divided into 1-3 charts, with a total of 8 charts describing the information dots and the chain of reasoning. Similarly, the Atlantic Storm dataset has 6 subplots and 13 charts. All datasets use irrelevant reports as noise to make the problem more challenging.
Statistics of the datasets and nodes created in DETs by the augmented architecture shown are shown in Table \ref{tab:dataset_stat}. We differentiate surface level ground truth text with implicit ground truth. An AI model's true capabilities in case of solving IA process can only be evaluated by comparing the implicit ground truth. For Crescent and Atlantic Storm, we isolate the implicit information by manually going through the chart and removing the document level dots.
\begin{table}[h]
    \centering
        \caption{Dataset Statistics}
        \label{tab:dataset_stat}
        \resizebox{0.95\columnwidth}{!}{%
        \begin{tabular}{@{}lcc@{}}
        \toprule
        \textbf{Dataset} & \textbf{\begin{tabular}[c]{@{}c@{}}\# Documents \\ (Relevant/Irrelevant)\end{tabular}} & \textbf{\begin{tabular}[c]{@{}c@{}}\# Nodes Generated \\ (Hypothesis/Evidential)\end{tabular}} \\ \midrule
        Crescent & 41 (25/16) & 45  (4/41) \\
        Atlantic Storm & 111 (65/46) & 128  (17/111) \\
        Manpad & 50  (21/29) & 50  (9/41) \\ \bottomrule
        \end{tabular}%
        }
\end{table}
\begin{figure}[h]
    \includegraphics[width=\columnwidth]{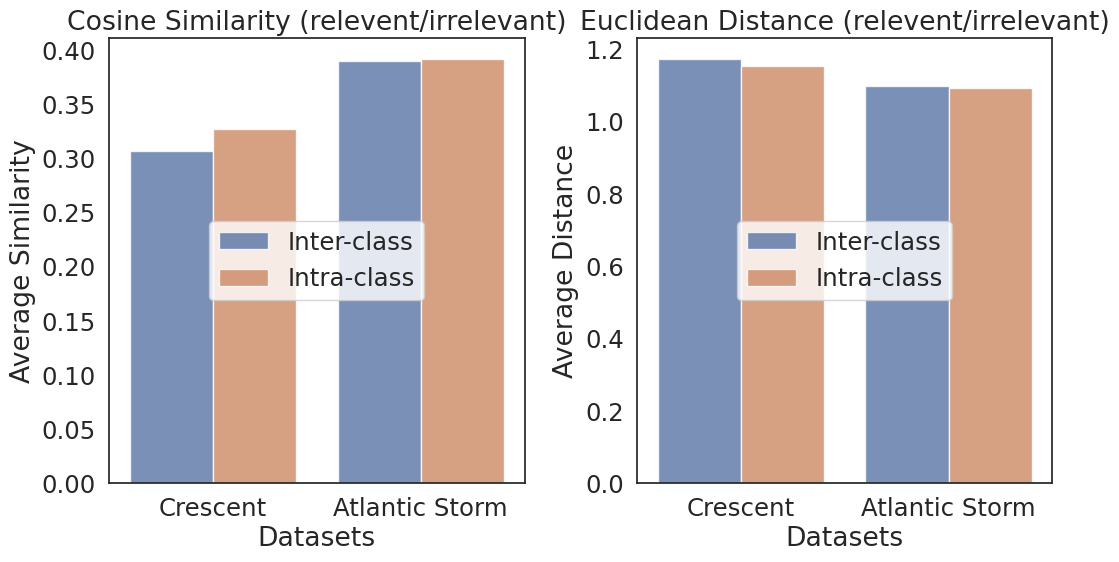}
    \caption{Intra and inter class distances and similarity for relevant and irrelevant documents in datasets}
    \label{fig:inter_intra}
\end{figure}
\subsection{Document-entity network and clustering}
We consider a document-entity network as the basic baseline, as demonstrated in the entity graph based approaches 
\cite{bier2009principles, hossain2012storytelling, wu2012start}. These works typically consider the story as an established connection between a set of entities by graph properties. The task of sense-making, finding out the bigger story, and narrative generation are still up to the analysts. Visual analytics tools were also developed to aid analysts with the sense-making task \cite{bier2006entity, gorg2012combining, stasko2007jigsaw}.

An objective of solving IA is to correctly isolate the irrelevant reports from the dataset. Traditional distance based methods, coupled with document embeddings, are often not enough to correctly classify the relevant reports. Fig.~\ref{fig:inter_intra} shows how the inter and intra class distance are almost equal for relevant and irrelevant reports for Crescent and Atlantic Storm datasets. To classify the relevant and irrelevant datasets, we used multiple clustering algorithms \cite{müllner2011modern, zhang1996birch} on the embedding vectors and report the maximum score for each dataset in Table \ref{tab:result_classification}.
\vspace{-0.2cm}
\subsection{Evaluation Metrics}
We report the F1 score for the relevant/irrelevant document classification accuracy in all three datasets. We also compare the ground truth narrative with model narrative by adopting traditional ML metrics, i.e., average ROUGE 1/2/L(ROUGE-Lsum) \cite{lin-2004-rouge}, and METEOR\cite{meteor}. To capture the quality of the narratives, we also employ GPT-4 \cite{achiam2023gpt} ratings. Recent works established LLMs as viable judge to evaluate the quality of open ended text generation \cite{Zheng2023JudgingLW, Fu2023GPTScoreEA, li2023generative_judge, huang2024empirical_judge, Hackl_2023} and to address the burdensome work of evaluation with human preferences \cite{wang2022selfinstruct, ouyang2022instructgpt, diao2023lmflow}. We identified three important qualities for the model responses, i.e., relevance, coverage, and thoughtfulness, and test GPT-4 ratings in respect to these qualities.
\vspace{-0.2cm}
\begin{table}[h]
    \centering
    \caption{Classification F-1 score on three datasets for LLM responses}
    \label{tab:result_classification}
    \resizebox{0.88\columnwidth}{!}{%
        \begin{tabular}{@{}lrrr@{}}
        \toprule
        \textbf{Method} & \multicolumn{1}{c}{\textbf{Crescent}} & \multicolumn{1}{c}{\textbf{Atlantic Storm}} & \multicolumn{1}{c}{\textbf{Manpad}} \\ \midrule
        Clustering & 0.65 & 0.72 & 0.56 \\
        Basic Prompt & n/a & n/a & n/a \\
        Basic Prompt (clustered) & 0.65 & 0.72 & 0.56 \\
        Augmented w/o data condensation & 0.80 & \textbf{0.80} & 0.64 \\ \midrule
        Full augmentation (Regular DET) & \textbf{0.81} & 0.78 & \textbf{0.65} \\ \bottomrule
        \end{tabular}%
    }
\end{table}
\begin{table*}[!t]
    \centering
    \caption{Evaluation of narrative generation.}
    \subcaption{ROUGE and METEOR scores on three datasets for GPT-3.5 (temperature=0.7)}
    \vspace{-0.2cm}
    \label{tab:result_narrative}
    \resizebox{0.95\textwidth}{!}{%
        \begin{tabular}{@{}lrrrrrrrrrrrr@{}}
        \toprule
        \textbf{}                           & \multicolumn{4}{c}{\textbf{Crescent}}                         & \multicolumn{4}{c}{\textbf{Atlantic Storm}}                   & \multicolumn{4}{c}{\textbf{Manpad}}                           \\ \midrule
        \textbf{Method}                     & METEOR        & R1            & R2            & RL            & METEOR        & R1            & R2            & RL            & METEOR        & R1            & R2            & RL            \\ \midrule
        Basic Prompt                        & 0.27          & 0.22          & 0.05          & 0.16          & 0.25          & 0.24          & 0.04          & 0.18          & 0.33          & 0.27          & 0.09          & 0.21          \\
        Basic Prompt (clustered)            & 0.26          & 0.18          & 0.05          & 0.14          & 0.21          & 0.21          & 0.03          & 0.15          & \textbf{0.35} & 0.35          & \textbf{0.12} & 0.26          \\
        DET (regular w/o data condensation) & 0.25          & 0.19          & 0.05          & 0.13          & 0.20          & 0.17          & 0.03          & 0.13          & 0.20          & 0.20          & 0.03          & 0.14          \\ \midrule
        DET (regular)                       & \textbf{0.29} & 0.18          & \textbf{0.07} & 0.13          & \textbf{0.34} & \textbf{0.29} & \textbf{0.10} & \textbf{0.21} & 0.27          & \textbf{0.36} & 0.07          & \textbf{0.28} \\
        DET (person-based)                  & 0.27          & \textbf{0.29} & \textbf{0.07} & \textbf{0.20} & 0.17          & 0.27          & 0.04          & 0.17          & 0.19          & 0.24          & 0.03          & 0.15          \\ \bottomrule
        \end{tabular}%
    }
    \smallskip
    \subcaption{GPT-4 Score on Likert chart (1-7) for difference quality of the narratives}
    \vspace{-0.2cm}
    \label{tab:gpt_v1}
    \resizebox{0.95\textwidth}{!}{%
    \begin{tabular}{@{}lccccccccc@{}}
    \toprule
    Dataset & \multicolumn{3}{c}{\textbf{Crescent}} & \multicolumn{3}{c}{\textbf{Atlantic Storm}} & \multicolumn{3}{c}{\textbf{Manpad}} \\ \midrule
    Methods & \textbf{Relevance} & \textbf{Coverage} & \textbf{Thoughtfulness} & \textbf{Relevance} & \textbf{Coverage} & \textbf{Thoughtfulness} & \textbf{Relevance} & \textbf{Coverage} & \textbf{Thoughtfulness} \\ \midrule
    Fully augmented & 5 & 3 & 5 & 2 & 2 & 5 & 1 & 1 & 4 \\
    Basic (sub-sampled) & 3 & 2 & 3 & 1 & 1 & 3 & 3 & 2.5 & 5 \\
    Basic (clustered) & 5 & 4 & 5 & 1 & 1 & 3 & 3 & 1 & 4 \\ \bottomrule
    \end{tabular}%
    }
    \smallskip
    \subcaption{ROUGE and METEOR scores on three datasets for different LLMs}
    \vspace{-0.2cm}
    \label{tab:diff_models}
    \resizebox{0.95\textwidth}{!}{%
    \begin{tabular}{@{}lrrrrrrrrrrrr@{}}
    \toprule
    \textbf{}      & \multicolumn{4}{c}{\textbf{Crescent}} & \multicolumn{4}{c}{\textbf{Atlantic Storm}} & \multicolumn{4}{c}{\textbf{Manpad}} \\ \midrule
    \textbf{Model} & METEOR    & R1      & R2     & RL     & METEOR     & R1       & R2       & RL       & METEOR   & R1     & R2     & RL     \\ \midrule
    GPT-4          & 0.25      & 0.23    & 0.07   & 0.15   & 0.23       & 0.23     & 0.04     & 0.16     & 0.28     & 0.31   & 0.07   & 0.19   \\
    Llama-2        & 0.20      & 0.19    & 0.03   & 0.11   & 0.21       & 0.21     & 0.03     & 0.15     & 0.28     & 0.36   & 0.08   & 0.26   \\
    Mistral-7B        & 0.26      & 0.23    & 0.07   & 0.16   & 0.25       & 0.20     & 0.04     & 0.15     & 0.34     & 0.37   & 0.11   & 0.26   \\
    Gemma-2        & 0.17      & 0.08    & 0.03   & 0.07   & 0.15       & 0.07     & 0.02     & 0.05     & 0.27     & 0.18   & 0.07   & 0.15   \\ \bottomrule
    \end{tabular}%
    }
\label{fig:all_fig}
\end{table*}
\begin{figure*}[h!]
\centering
    \begin{subfigure}[b]{\textwidth}
        \centering
        \includegraphics[width=\linewidth]{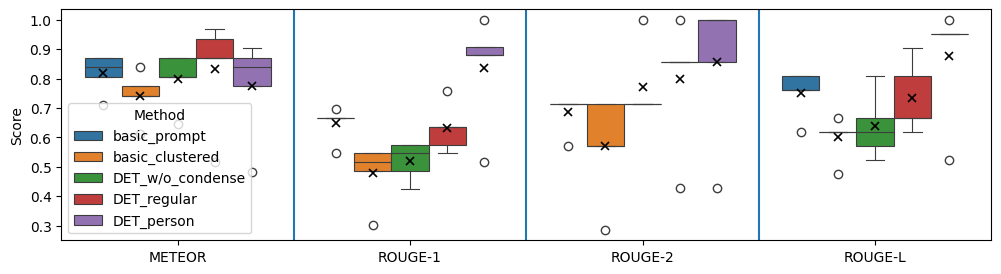}
        \vspace{-2\baselineskip}
        \caption{Crescent Dataset}
    \end{subfigure}
    \begin{subfigure}[b]{\textwidth}
        \centering
        \includegraphics[width=\linewidth]{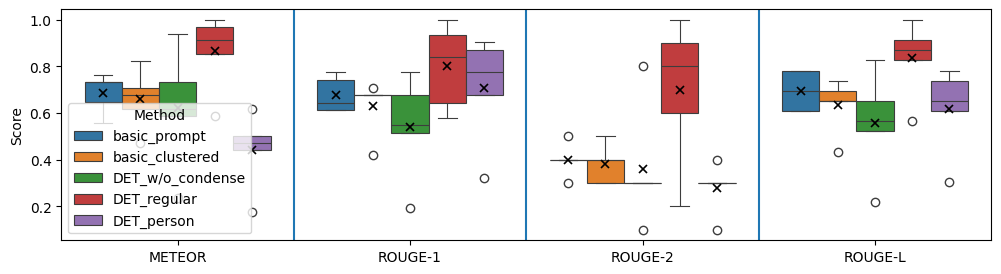}
        \vspace{-2\baselineskip}
        \caption{Atlantic Storm Dataset}
    \end{subfigure}
    \caption{Normalized metrics from the narrative for different methods with different temperatures [DET\_regular (and/or) DET\_person are higher for each metric, across different datasets]}
    \label{fig:boxplots}
\end{figure*}
\begin{figure*}[h]
\centering
\begin{subfigure}[b]{0.76\textwidth}
   \includegraphics[width=1\linewidth]{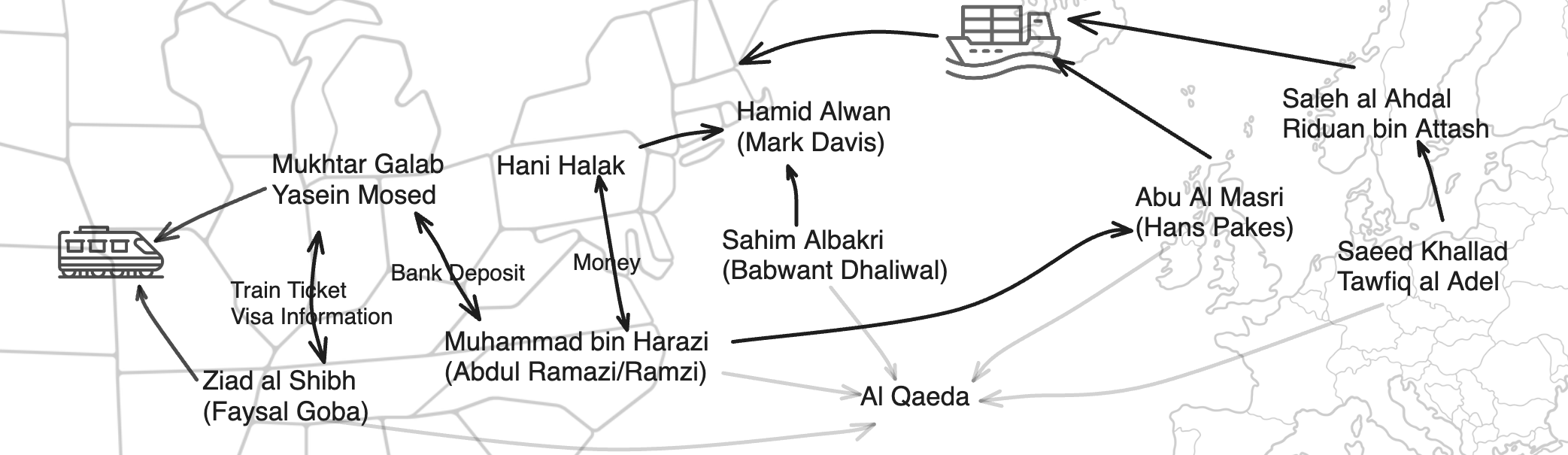}
   \caption{Sketch of the solution of Crescent dataset}
   \label{fig:crescent_case_study} 
\end{subfigure}
\begin{subfigure}[b]{0.95\textwidth}
   \includegraphics[width=1\linewidth]{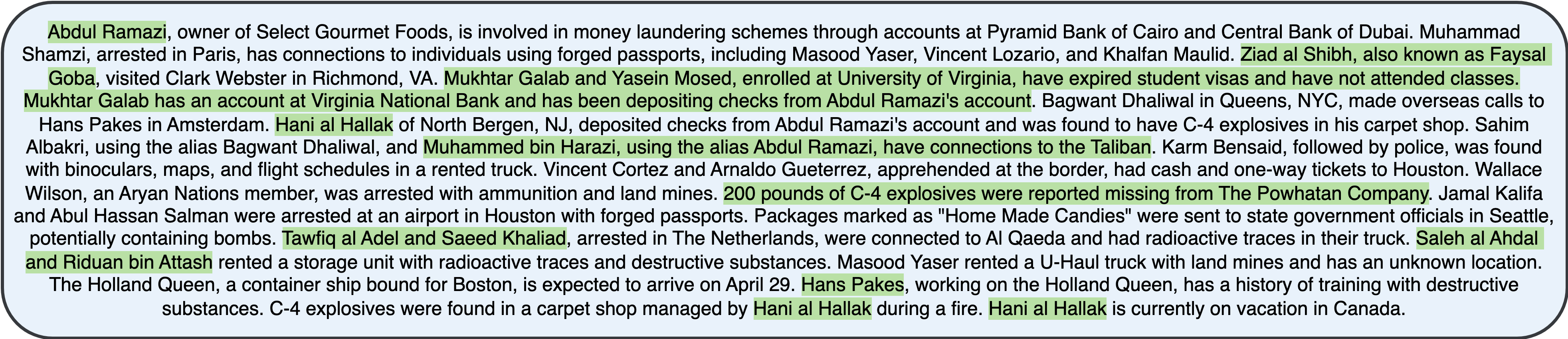}
   \caption{DETs augmented output for Crescent Dataset}
   \label{fig:crescent_output}
\end{subfigure}
\begin{subfigure}[b]{0.95\textwidth}
   \includegraphics[width=1\linewidth]{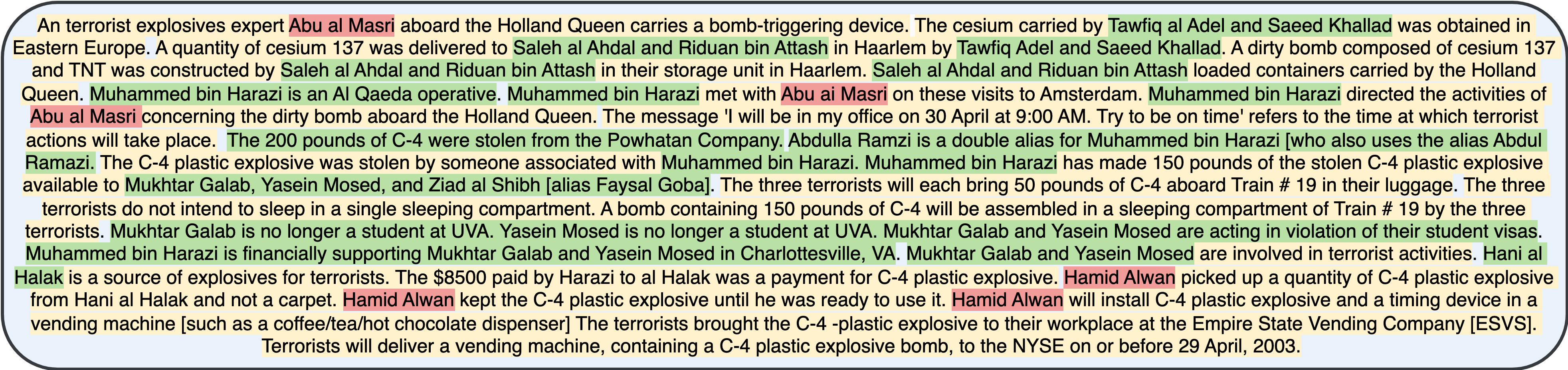}
   \caption{Ground truth narrative for Crescent Dataset [matched: green, and missed:(red: entities, yellow: implicit events)]}
   \label{fig:crescent_ground}
\end{subfigure}
\caption{Crescent dataset solution and narratives from LLMs with ground truth}
\label{fig:RQ1}
\end{figure*}
\section{Results}
\label{sec:results}
\vspace{-0.10cm}
Our experiments show that even with the augmented architecture, LLMs are unable to go beyond superficial reasoning. In this section, we expand on each RQ with the respective results.
\vspace{-0.1cm}
\subsection{RQ1 and RQ2: Can LLMs solve IA datasets on their own? Does augmentation help? If so what kind of augmentation helps?}
\label{ssec:RQ1_RQ2}
\vspace{-0.1cm}
To answer this set of RQs, we compare narratives from different experiments to the ground truth solutions. For each of Crescent and Atlantic Storm, we combined the ground truth of the charts into one report. For Manpad, we followed the rule-based grade to isolate relevant documents, with the entities extracted and desirable ones.

We first compare the classification accuracy for relevant documents. LLMs were only marginally better than clustering which suggests that the LLM-driven retrieval process is not upto mark, even though we had a two-step retrieval process.

Moreover, we also compare the narrative qualities against ground truth. The results for GPT-3.5 (Table \ref{tab:result_narrative}) show that the basic prompted LLM has unsatisfactory metrics against the ground truth. Because our ground truth is the implicit text, rather than the superficial text, it shows that LLM was unable to pick up and guess implicit stories from the documents. We also noticed a trend of subpar scores in the Atlantic Storm dataset which indicates that LLMs struggle with larger datasets. To answer if augmentations help, we plot the normalized metrics for different methods with multiple temperatures (0.2-1.5) in Fig. \ref{fig:boxplots}. DET (regular) and DET (person-based) show improved performance across metrics and datasets. We also observe the largest improvement with the augmentation occurs for Atlantic Storm, which indicates that the augmented version helps in better structuring the large sets of reports. In other datasets however, it only slightly outperforms basic strategy. Moreover, the augmented version's performance degraded once the data condensation feature was turned off, which suggests that, to build a structured architecture, summarization and cleaning the data helps tremendously.

Traditional evaluation metrics often fail to correctly quantify the open-ended generation for the desirable characteristics \cite{alabdulkarim2021storygen, Zheng2023JudgingLW}, so we also used GPT-4 to score the quality of the narratives. Across the qualities (i.e., relevance, coverage and thoughtfulness), we quantify the ratings in single Likert scale score (1-7) (Table \ref{tab:gpt_v1}). The trend confirms the narrative quality scores reported in Table \ref{tab:result_narrative}. However, in the next section, we expand on a potential pitfall of traditional evaluation systems.

\subsection{RQ3: What is the effect of temperature and allowable context size on the  skills of LLMs?}
\label{ssec:RQ3}
\vspace{-0.1cm}
We also investigate if it is possible to increase the creativity/analytical skills by utilizing the randomness and creativity focused prompts for language models.
For the basic setting, we tested two types of prompts with one containing system prompts about ``being creative and imaginative'' in reasoning.
For the augmented architecture, the quality of the narrative is affected by both temperature and allowable context window limit for each hypothesis dot. Rather than limiting the generation by setting the token length, we asked the model to curb generation within a certain word limit by prompt. 
We carried out parameter sweep for this setup on Crescent dataset and plot the results for combined ROUGE score and METEOR score (appendix).
Experiments show that for basic prompts, there are no trends in increasing temperature, for both type of prompts. For the augmented architecture in GPT models, we find two optimum setups for ROUGE and METEOR, i.e., (100 words and 0.7 temperature; 150 words and 0.5 temperature). Along with observation from RQ1 \& RQ2, it indicates that temperatures in the region of 0.5-1.0 work well. However, there is no improvement in higher temperatures, indicating that the reasoning capability is rather stagnant.
\vspace{-0.23cm}
\subsection{Use case studies}
\label{ssec:use_case}
Here we will qualitatively showcase the inability for the LLMs to look past the surface-level information. The goal of intelligence analysis is not just to summarize the reports, rather to find out the connections, along with back stories about the adversaries. We showcase the narratives from experiments for Crescent dataset in Fig. \ref{fig:crescent_output} and a high level sketch of the solution in Fig. \ref{fig:crescent_case_study}. Crescent dataset plot describes the plans for three synchronized attacks by Al Qaeda operatives, coordinated by Muhammad bin Harazi. Harazi sourced the funds and and organized operatives on three fronts, i) a container ship bound for Boston from Amsterdam, ii) Amtrak Train \#19 bound for Atlanta, and iii) the New York Stock Exchange. Ziad al Shibh, Sahim Albakri, Abu al Masri, Saeed Khallad, Tawfiq al Adel are the main individuals, cobwebbed by other operatives on each of three fronts. Reviewing the output against ground truth (Fig. \ref{fig:crescent_ground}), we find that LLMs lack on two fronts: i) failure to paint out an implicit story about the connections, ii) filtering out relevant and irrelevant documents. We also look at one particular subplot of Crescent dataset. The story is about a ship called ``Holland Queen'' and how it might have a potentially dangerous cargo onboard. Different versions of our LLMs only described the document-level entities and failed to make a guess that the ship might be carrying dangerous cargo.

\begin{figure*}[ht]
  \centering
    \includegraphics[width=0.9\textwidth]{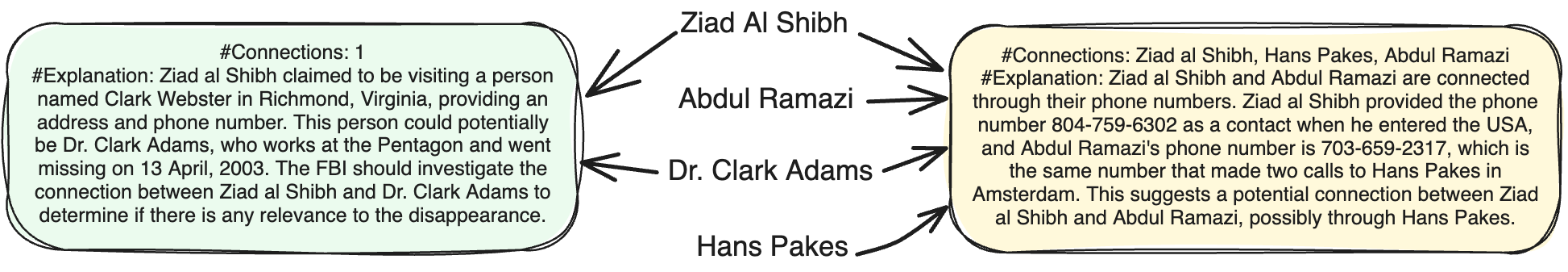}
    \caption{Small use case for imaginative reasoning: (Left) speculative in nature, combining two people based on the similarity of their name. (Right) failed to invoke the previous speculations about the names of the individuals and strayed away from making any credible speculation at all}
    \label{fig:speculation}
\end{figure*}
\vspace{-0.1cm}
\section{Recommendations}
\label{sec:recommendation}
We answer (\textbf{RQ4: Where can LLMs contribute in the 3-step process outlined in section \textsection 1}) in the form of recommendations from our findings. From the quantitative and qualitative experiments,
it is apparent that the generation of LLMs studied here struggle with in-depth analytical reasoning. Across all datasets, we see trends of good summarization, and less on analytical creativity or speculative/imaginative reasoning (not withstanding LLMs ability to hallucinate information). We briefly describe these points here:

\textbf{Steering speculative reasoning is hard:} LLMs are adept at summarizing documents into groups. However,  our repeated attempts with multiple types of prompts and parameter sweeps failed to invoke the capability required to properly speculate connections.
Complementary to this finding, we also experimented with a shorter use case with brief descriptions of four persons as shown in Fig. \ref{fig:speculation}. When asked about the connection between only two persons, LLMs were able to speculate based on the similarity of their names. However, with the addition of two more persons in the mix, LLMs failed to invoke the previous speculation. We also noticed that the position of the target entities mattered. Our finding is consistent with those of recent research, e.g., needle in the haystack \cite{needleinahaystack_2024}, lost in the middle \cite{liu2024lost} studies. Moreover, our findings suggest that, LLMs still struggle on a much smaller scale than previously thought of and this depends on both task complexity (i.e., number of entities in IA) and context length. We plan to address this in future work. 

\textbf{Larger is not necessarily better:} Despite GPT-4 being significantly larger, we find no discernible improvement as shown in Table \ref{tab:diff_models}. The results were consistent across different model families and indicates that the limitation in such reasoning is rather inherent to the LLMs in general. Notably, Gemma-2 scored lower than others, despite being a newer (and supposedly better) model which shows that it is important to establish such real-world and data-centric tasks for improving LLMs in general. As a ray of hope, preliminary tests on OpenAI's o1 model \cite{o1} (unavailable during submission) show substantial improvement which can be attributed to the additional chain-of-thought reasoning steps invoked during generation. This reinforces that additional reasoning augmentations can be helpful for IA tasks.

\textbf{LLMs are good organizers:} Even with the lack of in-depth analytical reasoning, LLMs were able to group related entities and events together, specially with smaller datasets. With large datasets, augmentation is needed to orchestrate some of the required grouping because of the context limitation and loss of attention \cite{levy2024same}.

\textbf{Orchestrating the evidence is a major challenge:} For data-centric analytical tasks like IA, it is hard for an LLM to take everything into context in a single prompt. Augmentation is necessary to organize the evidence. We propose a LLM-driven framework for these tasks. LLMs can also help in the search and retrieval process for such frameworks. Frameworks designed to show the immediate reasoning steps can also be helpful for the analysts to develop mental models.

\textbf{LLMs can generate good quality reports:} Although LLMs throughout our experiments lacked in-depth analytical/imaginative reasoning, they can produce good quality reports with the set of evidences. We suggest the use of LLMs as an interface before and after the data pipeline for such tasks.

\begin{table}[ht]
    \centering
    \caption{Pitfalls of traditional metrics in intelligence analysis (BERTScore/METEOR/ROUGE-1 scores)}
    \label{tab:pitfalls}
    \resizebox{\columnwidth}{!}{%
        \begin{tabular}{*{6}{c}}
        \toprule    
            & Crescent & Atlantic Storm & Manpad & Random text  \\ \midrule
          \multicolumn{1}{l}{Crescent} & X                 \\ 
          \multicolumn{1}{l}{Atlantic Storm} & \multicolumn{1}{r}{0.80/0.22/0.23} & X             \\ 
          \multicolumn{1}{l}{Manpad} & \multicolumn{1}{r}{0.80/0.09/0.23} & \multicolumn{1}{r}{0.79/0.05/0.12} & X             \\ 
          \multicolumn{1}{l}{Random text} & \multicolumn{1}{r}{0.77/0.06/0.12} & \multicolumn{1}{r}{0.77/0.03/0.06} & \multicolumn{1}{r}{0.79/0.12/0.19} & X         \\ 
          \multicolumn{1}{l}{Atlantic Storm} response & \multicolumn{1}{r}{0.80/0.20/0.23} & \multicolumn{1}{r}{0.81/0.21/0.34} & \multicolumn{1}{r}{0.81/0.19/0.16} & \multicolumn{1}{r}{0.79/0.15/0.07}    \\ \bottomrule
        \end{tabular}
    }
\end{table}

\textbf{Pitfalls of using traditional metrics:} We also noticed a significant limitation of traditional metrics, i.e., lexical similarity metrics (ROUGE, METEOR) and contextual embedding based metric (BERTScore) \cite{bertscore}. Due to the common themes, vocabulary, phrases, and content overlap across reports, these metrics exhibit unusually high scores, even across different datasets. This is particularly evident for the Atlantic Storm and Crescent datasets; the most notorious results being that of BERTScore, showing almost equal scores across different datasets. We report the findings in Table \ref{tab:pitfalls}. In light of these abnormalities, we suggest to use larger foundation models for automatic evaluations \cite{liu-etal-2023-g}.

\textbf{Data condensation is helpful:} Turning off data condensation had a noticeably negative impact on experimental results with DETs. This is because DETs rely on concise and precise text for efficient search, retrieval, and other operations. We recommend incorporating data condensation in any framework designed for data-centric applications, as it can significantly improve results.

\textbf{Classifying relevant/irrelevant documents:} One of the challenges for data-centric frameworks such as proposed here is the need for good search and retrieval capabilities; poor performance at this stage percolates downward to lackluster results, a theme well known from retrieval augmented generation (RAG) studies.

\textbf{Leveraging external knowledge sources:} We find some cases where it would be beneficial to have access to an external knowledge base or search engine. One such case arises in the Crescent dataset, where an information dot about reservations on AMTRAK Train \#19 was referred. Moreover, the word ``Crescent'' was also mentioned separately in later reports. Without access to external knowledge or the AMTRAK schedule, an LLM would probably not know that the Train \#19 is known as the Crescent Train \cite{noauthor_crescent_nodate}.

\textbf{Limitations:} As our experiments and augmentations are driven by prompt engineering~\cite{longpre2023flan, bach2022promptsource, wei2022chain} to enable LLMs to carry out text generation tasks, we acknowledge the potential variability in the wording of responses due to changes in prompts. To mitigate this aspect we empirically evaluated a range of prompts to establish a baseline for our studies.

\vspace{-0.1cm}
\section{Related Work}
From their original roots as text generators~\cite{achiam2023gpt,touvron2023llama}, LLMs have made their way into a range of downstream applications such as chat agents~\cite{park2023generative}, embodied game players~\cite{wang2023voyager}, and science solvers~\cite{romera2024mathematical}. The use of LLMs for IA remains under-explored. 

Pirolli and Card \cite{pirolli_sensemaking_2005}  were among the first to schematize the process of intelligence analysis using two interacting loops of activities: i) a foraging loop
with iterative search and extract of information
and ii) a sense-making loop
which involves developing a mental model with the information set at hand.
Similarly, Klein et al. \cite{klein_dataframe_2007}
described the IA process with two components and the interplay between them: i) data as the signals for the events and ii) the frame which explains the events. 
Sense-making is completed when frames are saturated with the data points.
Early approaches for aids were explored in different directions, e.g., model-based \cite{alonso2005model}, multi-agent based \cite{lindahl2007multi}, entity graph based \cite{bier2009principles, hossain2012storytelling, wu2012start}, and topic modeling based \cite{maiti2016interactive}. Various visual analytic systems were developed for entity graph based approaches~\cite{bier2006entity, gorg2012combining, stasko2007jigsaw} and for document level systems \cite{endert2014human, hossain2011helping}. A more recent approach to aid intelligence analysis process comes in the form of using large displays, leveraging virtual reality and immersive displays \cite{davidson2022exploring, lisle2023spaces}. All visual analytic approaches however leave the task of sense-making to the analysts. 

LLMs have long been augmented by external measures to enhance their intrinsic knowledge \cite{lewis2020retrieval, zhong2022training,wang2023selfknowledge,gao2023retrieval, shuster2022language}. LLMs have also been designed around external tools to perform real-world tasks \cite{lu2024chameleon_tool, schick2024toolformer, zhang2023graphtoolformer} and improve reasoning \cite{karpas2022mrkl, parisi2022talm}. There are also tool-based LLMs that can plan and use tools on their own \cite{qin2023toolllm, schick2024toolformer}. LLMs are also given external memory module to drive simulation~\cite{park2023generative}. Our study
aims to
uncover
whether LLMs are suitable as-is or with augmentations to support
complex tasks like intelligence analysis.
\section{Conclusion}
Strong analytical reasoning and the ability to synthesize information with an imaginative mindset are essential skills for intelligence analysts.
LLMs, with their generative strength, should have worked well to aid the analysts. However, as our study shows, effectively connecting the information dots across extensive and complex dossiers appears to be a research frontier. We believe this area holds significant potential for developing better primitives for assimilating knowledge and supporting sensemaking processes for analysts.

\bibliographystyle{IEEEtran}
\bibliography{IEEEfull}

\begin{thebibliography}{10}
\providecommand{\url}[1]{#1}
\csname url@samestyle\endcsname
\providecommand{\newblock}{\relax}
\providecommand{\bibinfo}[2]{#2}
\providecommand{\BIBentrySTDinterwordspacing}{\spaceskip=0pt\relax}
\providecommand{\BIBentryALTinterwordstretchfactor}{4}
\providecommand{\BIBentryALTinterwordspacing}{\spaceskip=\fontdimen2\font plus
\BIBentryALTinterwordstretchfactor\fontdimen3\font minus \fontdimen4\font\relax}
\providecommand{\BIBforeignlanguage}[2]{{%
\expandafter\ifx\csname l@#1\endcsname\relax
\typeout{** WARNING: IEEEtran.bst: No hyphenation pattern has been}%
\typeout{** loaded for the language `#1'. Using the pattern for}%
\typeout{** the default language instead.}%
\else
\language=\csname l@#1\endcsname
\fi
#2}}
\providecommand{\BIBdecl}{\relax}
\BIBdecl

\bibitem{kojima2022large}
T.~Kojima, S.~S. Gu, M.~Reid, Y.~Matsuo, and Y.~Iwasawa, ``Large language models are zero-shot reasoners,'' \emph{Advances in neural information processing systems}, vol.~35, pp. 22\,199--22\,213, 2022.

\bibitem{wei2022chain}
J.~Wei \emph{et~al.}, ``Chain-of-thought prompting elicits reasoning in large language models,'' \emph{Advances in neural information processing systems}, vol.~35, pp. 24\,824--24\,837, 2022.

\bibitem{bubeck2023sparks}
S.~Bubeck \emph{et~al.}, ``Sparks of artificial general intelligence: Early experiments with gpt-4,'' \emph{arXiv preprint arXiv:2303.12712}, 2023.

\bibitem{kumar2023large}
P.~Kumar, ``Large language models humanize technology,'' \emph{arXiv preprint arXiv:2305.05576}, 2023.

\bibitem{kaddour2023challenges}
J.~Kaddour \emph{et~al.}, ``Challenges and applications of large language models,'' \emph{arXiv preprint arXiv:2307.10169}, 2023.

\bibitem{medical_education}
C.~W. Safranek, A.~E. Sidamon-Eristoff, A.~Gilson, and D.~Chartash, ``The role of large language models in medical education: applications and implications,'' p. e50945, 2023.

\bibitem{dtg_survey}
M.~Sharma, A.~K. Gogineni, and N.~Ramakrishnan, ``Neural methods for data-to-text generation,'' \emph{ACM Transactions on Intelligent Systems and Technology}, 2024.

\bibitem{kang_characterizing}
Y.-a. Kang and J.~Stasko, ``Characterizing the intelligence analysis process: Informing visual analytics design through a longitudinal field study,'' in \emph{2011 IEEE Conference on Visual Analytics Science and Technology (VAST)}, 2011, pp. 21--30.

\bibitem{davidson2022exploring}
K.~Davidson, L.~Lisle, K.~Whitley, D.~A. Bowman, and C.~North, ``Exploring the evolution of sensemaking strategies in immersive space to think,'' \emph{IEEE transactions on visualization and computer graphics}, 2022.

\bibitem{touvron2023llama}
H.~Touvron \emph{et~al.}, ``Llama: Open and efficient foundation language models,'' \emph{arXiv preprint arXiv:2302.13971}, 2023.

\bibitem{touvron2023llama2}
------, ``Llama 2: Open foundation and fine-tuned chat models,'' \emph{arXiv preprint arXiv:2307.09288}, 2023.

\bibitem{ouyang2022instructgpt}
L.~Ouyang \emph{et~al.}, ``Training language models to follow instructions with human feedback,'' \emph{Advances in neural information processing systems}, vol.~35, pp. 27\,730--27\,744, 2022.

\bibitem{yao2024tree}
S.~Yao \emph{et~al.}, ``Tree of thoughts: Deliberate problem solving with large language models,'' \emph{Advances in Neural Information Processing Systems}, vol.~36, 2024.

\bibitem{besta2024graph}
M.~Besta \emph{et~al.}, ``Graph of thoughts: Solving elaborate problems with large language models,'' in \emph{Proceedings of the AAAI Conference on Artificial Intelligence}, vol.~38, no.~16, 2024, pp. 17\,682--17\,690.

\bibitem{levy2024same}
M.~Levy, A.~Jacoby, and Y.~Goldberg, ``Same task, more tokens: the impact of input length on the reasoning performance of large language models,'' \emph{arXiv preprint arXiv:2402.14848}, 2024.

\bibitem{park2023generative}
J.~S. Park \emph{et~al.}, ``Generative agents: Interactive simulacra of human behavior,'' in \emph{Proceedings of the 36th Annual ACM Symposium on User Interface Software and Technology}, 2023, pp. 1--22.

\bibitem{muennighoff2022mteb}
N.~Muennighoff, N.~Tazi, L.~Magne, and N.~Reimers, ``Mteb: Massive text embedding benchmark,'' \emph{arXiv preprint arXiv:2210.07316}, 2022.

\bibitem{huang2020embedding}
J.-T. Huang \emph{et~al.}, ``Embedding-based retrieval in facebook search,'' in \emph{Proceedings of the 26th ACM SIGKDD International Conference on Knowledge Discovery \& Data Mining}, 2020, pp. 2553--2561.

\bibitem{su2023embedder}
H.~Su \emph{et~al.}, ``One embedder, any task: Instruction-finetuned text embeddings,'' 2023.

\bibitem{jiang2023mistral7b}
A.~Q. Jiang \emph{et~al.}, ``Mistral 7b,'' \emph{arXiv preprint arXiv:2310.06825}, 2023.

\bibitem{gemma}
G.~Team \emph{et~al.}, ``Gemma: Open models based on gemini research and technology,'' \emph{arXiv preprint arXiv:2403.08295}, 2024.

\bibitem{wu2012start}
H.~Wu \emph{et~al.}, ``Where do i start? algorithmic strategies to guide intelligence analysts,'' in \emph{Proceedings of the ACM SIGKDD Workshop on Intelligence and Security Informatics}, 2012, pp. 1--8.

\bibitem{bier2009principles}
E.~A. Bier, S.~K. Card, and J.~W. Bodnar, ``Principles and tools for collaborative entity-based intelligence analysis,'' \emph{IEEE transactions on visualization and computer graphics}, vol.~16, no.~2, pp. 178--191, 2009.

\bibitem{hossain2012storytelling}
M.~S. Hossain, P.~Butler, A.~P. Boedihardjo, and N.~Ramakrishnan, ``Storytelling in entity networks to support intelligence analysts,'' in \emph{Proceedings of the 18th ACM SIGKDD international conference on Knowledge discovery and data mining}, 2012, pp. 1375--1383.

\bibitem{bier2006entity}
E.~A. Bier, E.~W. Ishak, and E.~Chi, ``Entity workspace: An evidence file that aids memory, inference, and reading,'' in \emph{Intelligence and Security Informatics}, S.~Mehrotra, D.~D. Zeng, H.~Chen, B.~Thuraisingham, and F.-Y. Wang, Eds., 2006, pp. 466--472.

\bibitem{gorg2012combining}
C.~G{\"o}rg \emph{et~al.}, ``Combining computational analyses and interactive visualization for document exploration and sensemaking in jigsaw,'' \emph{IEEE transactions on Visualization and Computer Graphics}, vol.~19, no.~10, pp. 1646--1663, 2012.

\bibitem{stasko2007jigsaw}
J.~Stasko, C.~Gorg, Z.~Liu, and K.~Singhal, ``Jigsaw: supporting investigative analysis through interactive visualization,'' in \emph{2007 IEEE Symposium on Visual Analytics Science and Technology}.\hskip 1em plus 0.5em minus 0.4em\relax IEEE, 2007, pp. 131--138.

\bibitem{müllner2011modern}
D.~Müllner, ``Modern hierarchical, agglomerative clustering algorithms,'' 2011.

\bibitem{zhang1996birch}
T.~Zhang, R.~Ramakrishnan, and M.~Livny, ``Birch: an efficient data clustering method for very large databases,'' \emph{ACM sigmod record}, vol.~25, no.~2, pp. 103--114, 1996.

\bibitem{lin-2004-rouge}
C.-Y. Lin, ``Rouge: A package for automatic evaluation of summaries,'' in \emph{Text summarization branches out}, 2004, pp. 74--81.

\bibitem{meteor}
S.~Banerjee and A.~Lavie, ``Meteor: an automatic metric for mt evaluation with high levels of correlation with human judgments,'' \emph{Proceedings of ACL-WMT}, pp. 65--72, 2004.

\bibitem{achiam2023gpt}
J.~Achiam \emph{et~al.}, ``Gpt-4 technical report,'' \emph{arXiv preprint arXiv:2303.08774}, 2023.

\bibitem{Zheng2023JudgingLW}
L.~Zheng \emph{et~al.}, ``Judging llm-as-a-judge with mt-bench and chatbot arena,'' \emph{Advances in Neural Information Processing Systems}, vol.~36, pp. 46\,595--46\,623, 2023.

\bibitem{Fu2023GPTScoreEA}
J.~Fu, S.-K. Ng, Z.~Jiang, and P.~Liu, ``Gptscore: Evaluate as you desire,'' \emph{arXiv preprint arXiv:2302.04166}, 2023.

\bibitem{li2023generative_judge}
J.~Li \emph{et~al.}, ``Generative judge for evaluating alignment,'' \emph{arXiv preprint arXiv:2310.05470}, 2023.

\bibitem{huang2024empirical_judge}
H.~Huang, Y.~Qu, J.~Liu, M.~Yang, and T.~Zhao, ``An empirical study of llm-as-a-judge for llm evaluation: Fine-tuned judge models are task-specific classifiers,'' \emph{arXiv preprint arXiv:2403.02839}, 2024.

\bibitem{Hackl_2023}
V.~Hackl, A.~E. M{\"u}ller, M.~Granitzer, and M.~Sailer, ``Is gpt-4 a reliable rater? evaluating consistency in gpt-4's text ratings,'' in \emph{Frontiers in Education}, vol.~8.\hskip 1em plus 0.5em minus 0.4em\relax Frontiers Media SA, 2023, p. 1272229.

\bibitem{wang2022selfinstruct}
Y.~Wang \emph{et~al.}, ``Self-instruct: Aligning language models with self-generated instructions,'' \emph{arXiv preprint arXiv:2212.10560}, 2022.

\bibitem{diao2023lmflow}
S.~Diao \emph{et~al.}, ``Lmflow: An extensible toolkit for finetuning and inference of large foundation models,'' \emph{arXiv preprint arXiv:2306.12420}, 2023.

\bibitem{alabdulkarim2021storygen}
A.~Alabdulkarim, S.~Li, and X.~Peng, ``Automatic story generation: Challenges and attempts,'' \emph{arXiv preprint arXiv:2102.12634}, 2021.

\bibitem{needleinahaystack_2024}
\BIBentryALTinterwordspacing
gkamradt, ``gkamradt/{LLMTest}\_needleinahaystack,'' Jul. 2024, original-date: 2023-11-11T00:50:02Z. [Online]. Available: \url{https://github.com/gkamradt/LLMTest_NeedleInAHaystack}
\BIBentrySTDinterwordspacing

\bibitem{liu2024lost}
N.~F. Liu \emph{et~al.}, ``Lost in the middle: How language models use long contexts,'' \emph{Transactions of the Association for Computational Linguistics}, vol.~12, pp. 157--173, 2024.

\bibitem{o1}
\BIBentryALTinterwordspacing
``\BIBforeignlanguage{en-US}{Learning to {Reason} with {LLMs}}.'' [Online]. Available: \url{https://openai.com/index/learning-to-reason-with-llms/}
\BIBentrySTDinterwordspacing

\bibitem{bertscore}
T.~Zhang, V.~Kishore, F.~Wu, K.~Q. Weinberger, and Y.~Artzi, ``Bertscore: Evaluating text generation with bert,'' \emph{arXiv preprint arXiv:1904.09675}, 2019.

\bibitem{liu-etal-2023-g}
Y.~Liu \emph{et~al.}, ``G-eval: Nlg evaluation using gpt-4 with better human alignment,'' \emph{arXiv preprint arXiv:2303.16634}, 2023.

\bibitem{noauthor_crescent_nodate}
\BIBentryALTinterwordspacing
``Crescent {Train} {New} {York}, {Atlanta}, {New} {Orleans} {\textbar} {Amtrak}.'' [Online]. Available: \url{https://www.amtrak.com/crescent-train}
\BIBentrySTDinterwordspacing

\bibitem{longpre2023flan}
S.~Longpre \emph{et~al.}, ``The flan collection: Designing data and methods for effective instruction tuning,'' in \emph{International Conference on Machine Learning}.\hskip 1em plus 0.5em minus 0.4em\relax PMLR, 2023, pp. 22\,631--22\,648.

\bibitem{bach2022promptsource}
S.~H. Bach \emph{et~al.}, ``Promptsource: An integrated development environment and repository for natural language prompts,'' \emph{arXiv preprint arXiv:2202.01279}, 2022.

\bibitem{wang2023voyager}
G.~Wang \emph{et~al.}, ``Voyager: An open-ended embodied agent with large language models,'' \emph{arXiv preprint arXiv:2305.16291}, 2023.

\bibitem{romera2024mathematical}
B.~Romera-Paredes \emph{et~al.}, ``Mathematical discoveries from program search with large language models,'' \emph{Nature}, vol. 625, no. 7995, pp. 468--475, 2024.

\bibitem{pirolli_sensemaking_2005}
P.~Pirolli and S.~Card, \emph{The sensemaking process and leverage points for analyst technology as identified through cognitive task analysis}, Jan. 2005.

\bibitem{klein_dataframe_2007}
G.~Klein, J.~K. Phillips, E.~L. Rall, and D.~A. Peluso, ``A {Data}–{Frame} {Theory} of {Sensemaking},'' in \emph{Expertise {Out} of {Context}}.\hskip 1em plus 0.5em minus 0.4em\relax Psychology Press, 2007, num Pages: 43.

\bibitem{alonso2005model}
R.~Alonso and H.~Li, ``Model-guided information discovery for intelligence analysis,'' in \emph{Proceedings of the 14th ACM international conference on Information and knowledge management}, 2005, pp. 269--270.

\bibitem{lindahl2007multi}
E.~Lindahl, S.~O'Hara, and Q.~Zhu, ``A multi-agent system of evidential reasoning for intelligence analyses,'' in \emph{Proceedings of the 6th international joint conference on Autonomous agents and multiagent systems}, 2007, pp. 1--6.

\bibitem{maiti2016interactive}
D.~Maiti, M.~R. Islam, and N.~Ramakrishnan, ``Interactive storytelling over document collections,'' \emph{arXiv preprint arXiv:1602.06566}, 2016.

\bibitem{endert2014human}
A.~Endert \emph{et~al.}, ``The human is the loop: new directions for visual analytics,'' \emph{Journal of intelligent information systems}, vol.~43, pp. 411--435, 2014.

\bibitem{hossain2011helping}
M.~S. Hossain, C.~Andrews, N.~Ramakrishnan, and C.~North, ``Helping intelligence analysts make connections,'' in \emph{Workshops at the Twenty-Fifth AAAI Conference on Artificial Intelligence}, 2011.

\bibitem{lisle2023spaces}
L.~Lisle \emph{et~al.}, ``Spaces to think: A comparison of small, large, and immersive displays for the sensemaking process,'' in \emph{2023 IEEE International Symposium on Mixed and Augmented Reality (ISMAR)}.\hskip 1em plus 0.5em minus 0.4em\relax IEEE, 2023, pp. 1084--1093.

\bibitem{lewis2020retrieval}
P.~Lewis \emph{et~al.}, ``Retrieval-augmented generation for knowledge-intensive nlp tasks,'' \emph{Advances in Neural Information Processing Systems}, vol.~33, pp. 9459--9474, 2020.

\bibitem{zhong2022training}
Z.~Zhong, T.~Lei, and D.~Chen, ``Training language models with memory augmentation,'' \emph{arXiv preprint arXiv:2205.12674}, 2022.

\bibitem{wang2023selfknowledge}
Y.~Wang, P.~Li, M.~Sun, and Y.~Liu, ``Self-knowledge guided retrieval augmentation for large language models,'' 2023.

\bibitem{gao2023retrieval}
Y.~Gao \emph{et~al.}, ``Retrieval-augmented generation for large language models: A survey,'' \emph{arXiv preprint arXiv:2312.10997}, 2023.

\bibitem{shuster2022language}
K.~Shuster \emph{et~al.}, ``Language models that seek for knowledge: Modular search \& generation for dialogue and prompt completion,'' \emph{arXiv preprint arXiv:2203.13224}, 2022.

\bibitem{lu2024chameleon_tool}
P.~Lu \emph{et~al.}, ``Chameleon: Plug-and-play compositional reasoning with large language models,'' \emph{Advances in Neural Information Processing Systems}, vol.~36, 2024.

\bibitem{schick2024toolformer}
T.~Schick \emph{et~al.}, ``Toolformer: Language models can teach themselves to use tools,'' \emph{Advances in Neural Information Processing Systems}, vol.~36, 2024.

\bibitem{zhang2023graphtoolformer}
J.~Zhang, ``Graph-toolformer: To empower llms with graph reasoning ability via prompt augmented by chatgpt,'' \emph{arXiv preprint arXiv:2304.11116}, 2023.

\bibitem{karpas2022mrkl}
E.~Karpas \emph{et~al.}, ``Mrkl systems: A modular, neuro-symbolic architecture that combines large language models, external knowledge sources and discrete reasoning,'' \emph{arXiv preprint arXiv:2205.00445}, 2022.

\bibitem{parisi2022talm}
A.~Parisi, Y.~Zhao, and N.~Fiedel, ``Talm: Tool augmented language models,'' \emph{arXiv preprint arXiv:2205.12255}, 2022.

\bibitem{qin2023toolllm}
Y.~Qin \emph{et~al.}, ``Toolllm: Facilitating large language models to master 16000+ real-world apis,'' \emph{arXiv preprint arXiv:2307.16789}, 2023.

\end{thebibliography}
\appendix
\label{app:temp}
\begin{figure}[H]
    \centering
        \begin{subfigure}[t]{0.45\columnwidth}
        \includegraphics[width=\linewidth]{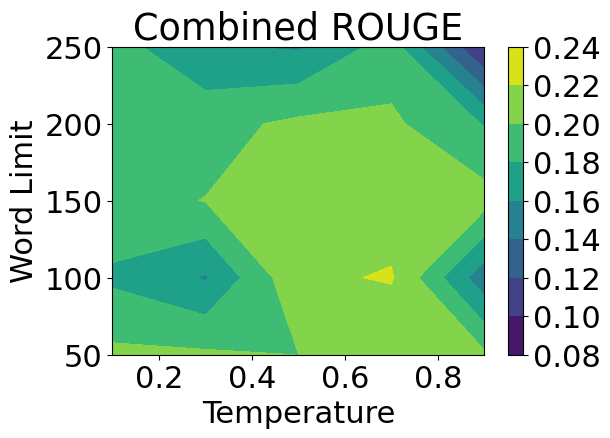}
        \label{fig:a}
    \end{subfigure}
    \hfill
    \begin{subfigure}[t]{0.45\columnwidth}
        \includegraphics[width=\linewidth]{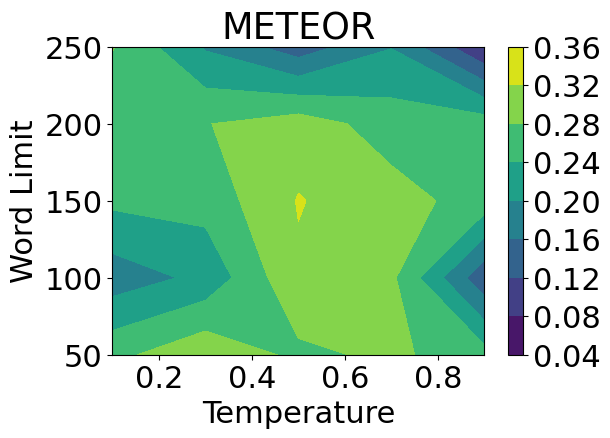}
        \label{fig:b}
    \end{subfigure}%
    
    \begin{subfigure}[b]{0.45\columnwidth}
          \includegraphics[width=\linewidth]{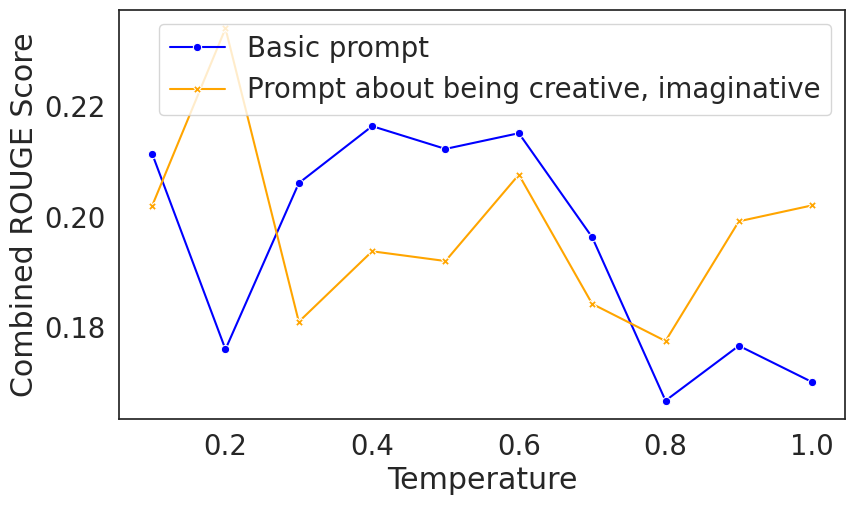}
        \label{fig:c}
    \end{subfigure}
    \hfill
    \begin{subfigure}[b]{0.45\columnwidth}
        \includegraphics[width=\linewidth]{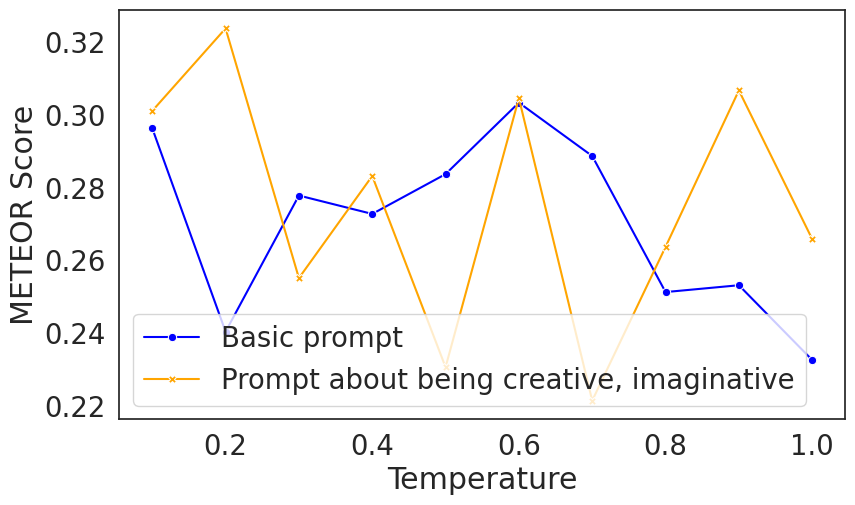}
        \label{fig:d}
    \end{subfigure}%
    \caption{Effect of temperature, different prompt style, allowable context length on LLM responses}
    \label{fig:temp_effect}
\end{figure}

\end{document}